\documentclass[letterpaper, 10 pt, conference]{ieeeconf}
\IEEEoverridecommandlockouts
\usepackage{graphicx} \graphicspath{{figures/}}
\usepackage{times,latexsym}
\usepackage{amsmath,amssymb}
\usepackage{algorithmic}
\usepackage[linesnumbered,ruled,vlined]{algorithm2e}
\usepackage{acronym}
\usepackage{balance}
\usepackage{xspace}
\usepackage{setspace}
\usepackage[skip=3pt,font=small]{subcaption}
\usepackage[skip=3pt,font=small]{caption}
\usepackage[dvipsnames,svgnames,x11names]{xcolor}
\usepackage{booktabs,tabularx,colortbl,multirow,array,makecell}
\makeatletter
\let\NAT@parse\undefined
\makeatother
\usepackage[
    colorlinks=true,
    linkcolor=black,
]{hyperref}
\usepackage[capitalise,nameinlink]{cleveref}
\usepackage{overpic,wrapfig}
\usepackage{cite}
\usepackage[misc]{ifsym}
\usepackage{threeparttable}
\usepackage{tablefootnote}
\usepackage{adjustbox}
\usepackage{placeins}

\def\method{\textbf{\textsc{Fiber}}\xspace}
\def\bfm{\textbf{\textsc{AH-BFM}}\xspace}

\newcommand{\pibfm}{\pi_\mathrm{BFM}}

\usepackage{tabularx,booktabs,ragged2e}
\newcolumntype{J}{>{\RaggedRight\arraybackslash}X}

\definecolor{WeiColor}{rgb}{1,0.33,0.64}
\definecolor{myDarkBlue}{RGB}{55,81,139}
\definecolor{LimeGreen}{RGB}{158,186,217}
\definecolor{myLightGreen}{RGB}{126,171,85}
\definecolor{myRed}{RGB}{176,36,24}

\makeatletter
\DeclareRobustCommand\onedot{\futurelet\@let@token\@onedot}
\def\@onedot{\ifx\@let@token.\else.\null\fi\xspace}

\makeatother

\crefname{algorithm}{Alg.}{Algs.}
\Crefname{algocf}{Alg.}{Algs.}
\crefname{section}{Sec.}{Secs.}
\Crefname{section}{Section}{Sections}
\crefname{table}{Tab.}{Tabs.}
\Crefname{table}{TABLE}{TABLES}
\crefname{figure}{Fig.}{Fig.}
\Crefname{figure}{Figure}{Figure}

\acrodef{llm}[LLM]{Large Language Model}
\acrodef{rl}[RL]{reinforcement learning}
\acrodef{bfm}[BFM]{Behavior Foundation Model}
\acrodef{fb}[FB]{Forward-Backward}
\acrodef{mpc}[MPC]{Model Predictive Control}
\acrodef{mlp}[MLP]{Multi-Layer Perceptron}
\acrodef{mdp}[MDP]{Markov Decision Process}
\acrodef{ppo}[PPO]{Proximal Policy Optimization}
\acrodef{slerp}[SLERP]{Spherical Linear Interpolation}
\acrodef{wbc}[WBC]{Whole-Body Control}
\acrodef{lpp}[LPP]{Latent Policy Projection}
\acrodef{rtc}[RTC]{Real Time Chunking}
\acrodef{sf}[SF]{Successor-Feature}

\title{\LARGE \bf Action-Aligned Latent Planning for Versatile \\Humanoid Loco-Manipulation}
\author{
    Yutang Lin$^{1}$ \qquad
    Guanya Shi$^{1,2,\dagger}$ \qquad
    C. Karen Liu$^{1,3,\dagger}$
    \thanks{Project website: \url{https://fiberaa.github.io/}.}
    \\[3pt]
    {\small
        $^{1}$Amazon FAR \quad
        $^{2}$Carnegie Mellon University \quad
        $^{3}$Stanford University
    }
    \\[3pt]
    {\footnotesize
        $^\dagger$Amazon FAR Team Co-Leads
    }
}

\let\oldtwocolumn\twocolumn
\renewcommand\twocolumn[1][]{%
    \oldtwocolumn[{#1}{
    \vspace{-1.4em}
    \begin{center}
            \includegraphics[width=\textwidth]{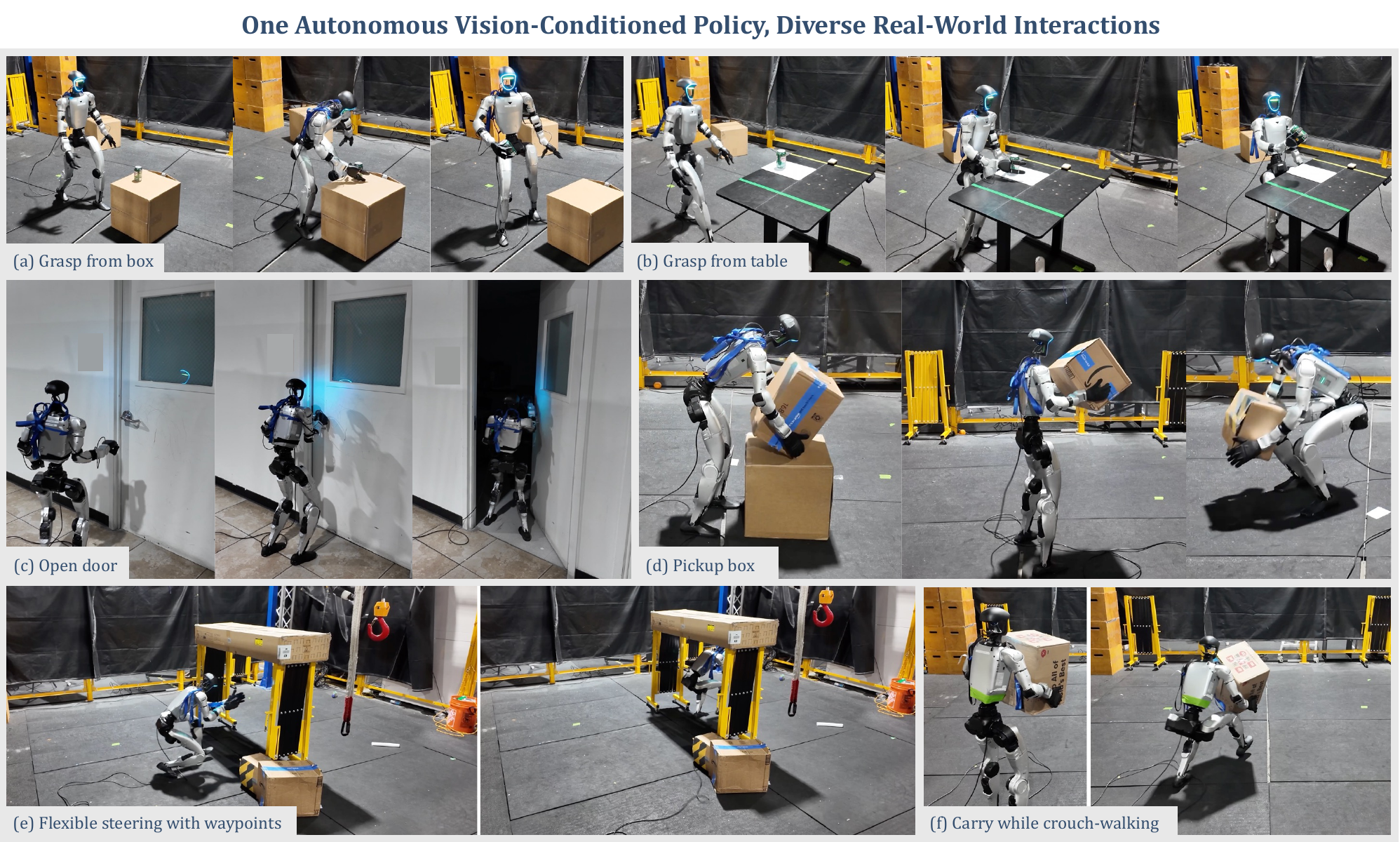}
            \captionsetup[figure]{hypcap=false}
            \captionof{figure}{
            A single autonomous perceptive latent planner enables diverse real-world
            interactions, including object grasping in (a,b), door opening
            in (c), whole-body box pickup in (d), waypoint-conditioned steering in (e), and
            crouched locomotion while carrying in (f).
            }
            \label{fig:teaser}
        \end{center}
    }]
}

\begin{document}
\maketitle
\thispagestyle{empty}
\pagestyle{empty}
\bstctlcite{IEEEexample:BSTcontrol}

\begin{abstract}
% [Intro]
Learning a single perceptive policy for diverse humanoid loco-manipulation requires unifying high-level task decisions without losing the state-dependent physical responses that make each interaction succeed. 
% [Background]
This makes the representation between high-level planning and whole-body control critical: kinematic or contact targets provide compact abstractions but do not directly encode interaction-specific control responses, while directly predicting low-level actions requires the planner to model high-frequency closed-loop feedback.
% [Propose]
We propose \method, a framework that uses the executable latent space of a frozen \ac{bfm} as a shared planning representation for heterogeneous interaction skills.
Our \ac{lpp} maps independently acquired task-specific policies into this representation by identifying latent commands whose induced actions match their local control responses, enabling lower-rate planning while retaining the BFM’s high-frequency state feedback.
After projection, we aggregate closed-loop latent rollouts from all skills and
train a single perceptive flow planner entirely offline.
We further introduce \bfm, an adaptive-horizon \ac{bfm} trained with
multi-$\gamma$ \ac{fb} learning to support both precise control
and temporally extended whole-body behavior.
% [Experiment]
We evaluate \method on box pickup, door opening, and object grasping in
simulation and on a real Unitree G1 humanoid.
Across 128 diverse variations per task in simulation, \method achieves
$42.0\%$ mean success, more than tripling projected latent replay ($11.6\%$)
and substantially outperforming kinematic planning ($3.5\%$).
Real-world experiments further demonstrate sim-to-real transfer, skill
chaining, and conditional steering through the same unified planner.
\end{abstract}

\acresetall

\section{Introduction}

% [Background: from task-specific skills to a unified policy]
Humanoid loco-manipulation requires both deciding \emph{what} interaction to
perform from visual and semantic context and determining \emph{how} to realize
it through balance, contact, and coordinated whole-body control.
Recent work has achieved strong task-specific interaction skills, including
object manipulation, articulated-object interaction, and object
transport~\cite{he2025viral,xue2026doorman}, often using task-specific
objectives, observations, and training procedures.
Scaling these successes to a versatile policy requires unifying diverse
task-level decisions without losing the physical responses that make each
skill succeed.

% [Hierarchical control and the representation challenge]
A common strategy separates task-level decision making from whole-body control.
A high-level policy predicts structured targets such as body motion,
end-effector poses, or contacts, while a low-level controller realizes them on
the robot.
However, the same target may require different actions under different contacts,
object states, or disturbances, so such representations do not directly capture
the state-dependent control responses required for interaction.
Predicting low-level actions avoids this abstraction, but offline learning then
requires broad coverage of closed-loop corrective responses, which is difficult
for contact-rich interaction.
This exposes a central challenge for humanoid planning: \textbf{\textit{how can we abstract low-level control complexity without abstracting away the physical responses that make interaction succeed?}}

% [Learned behavioral interface]
A learned behavioral representation offers a way to resolve this tension.
We build on a \ac{bfm} that exposes an executable behavioral latent space.
Given the current robot state and a latent command, \ac{bfm} produces low-level actions at high frequency.
A latent therefore indexes a state-dependent control policy rather than merely a desired motion, allowing high-level planning to operate over compact behavioral commands while the \ac{bfm} retains high-frequency closed-loop physical feedback.

% [How to populate the interface with diverse interaction capabilities]
Using such an interface for versatile loco-manipulation still requires acquiring the interaction capabilities that the planner should invoke.
One approach is to collect demonstrations for the desired interactions through a common data-acquisition pipeline~\cite{li2026omega0,zheng2026motionwam}, but obtaining broad coverage of contact-rich and corrective behavior is expensive.
Another approach is to first train task-specific interaction policies and distill them into a generalist.
On-policy multi-expert distillation can improve closed-loop coverage by querying these policies at states visited by the generalist, but requires all policies and their environments to remain available during integration.
We instead consider a decoupled setting: each interaction policy can be acquired independently, and only afterward incorporated into the shared behavioral interface.

% [Key insight: robot actions provide the common grounding]
Under this decoupled setting, different interaction policies need not share
their observations, objectives, or task representations.
They nevertheless share one quantity that directly describes their physical
realization: the actions applied to the robot.
A latent command from the frozen \ac{bfm} is grounded in the same quantity,
because it also produces robot actions conditioned on the current robot state.
This common action space provides a direct way to relate an independently
acquired interaction policy to the \ac{bfm} latent representation.

% [FIBER and action-aligned latent policy projection]
Based on this insight, we propose \method, which represents independently
acquired interaction policies, or experts, in the executable latent space of a frozen \ac{bfm}.
At the core of \method is \ac{lpp}, which identifies latent commands by matching
the actions induced by the \ac{bfm} to an expert's local state-dependent control
responses.
The resulting projected policy executes through the frozen \ac{bfm}, preserving
high-frequency state feedback while expressing heterogeneous interaction skills
through a common lower-rate representation.

% [Offline unified planning]
Because each expert is projected independently, the experts need not share an
acquisition procedure or remain jointly available during subsequent training.
After projection, we roll out the projected policies and collect their latent
commands together with the perceptive and proprioceptive observations available
to the final policy.
These closed-loop latent rollouts are aggregated across skills to train a single
perceptive planner entirely offline.
At deployment, the planner predicts future latent commands from onboard
observations and a high-level steering condition, while the frozen \ac{bfm}
provides the state-dependent whole-body responses required to execute them.

% [Adaptive-horizon BFM]
Finally, projection and planning impose different temporal requirements on the latent representation.
Matching an expert's control response favors precise short-horizon
behavior, whereas lower-rate planning benefits from stable commands that preserve behavior over longer horizons.
We therefore introduce \bfm, an adaptive-horizon \ac{bfm} trained with
multi-$\gamma$ \ac{fb} learning to support both control fidelity
and temporally extended behavioral structure.

% [Evaluation]
We evaluate \method on three heterogeneous interaction tasks---box pickup, door
opening, and object grasping-in simulation and on a real Unitree G1.
In simulation, \ac{lpp} largely preserves expert performance on box pickup and
door opening near the interaction regime.
The unified perceptive planner achieves $42.0\%$ mean success, compared with
$11.6\%$ for fixed projected latent replay and $3.5\%$ for kinematic planning.
Real-world experiments further demonstrate direct sim-to-real transfer,
interaction chaining, and waypoint-conditioned steering through the same
planner.

Our contributions are threefold:
\begin{itemize}
    \item
    \textbf{A shared executable representation for humanoid interaction
    planning.}
    We use the latent space of a frozen \ac{bfm} as a behavioral interface that
    supports lower-rate high-level planning while retaining state-dependent
    interaction realization responses.

    \item
    \textbf{Action-aligned latent policy projection.}
    We introduce \ac{lpp}, which maps independently acquired interaction
    experts into the shared \ac{bfm} representation by aligning their local
    control responses in robot action space, decoupling skill acquisition from
    subsequent integration.

    \item 
    \textbf{Adaptive-horizon BFM and SF-guided behavior reuse.}
    We introduce multi-$\gamma$ \ac{fb} learning and \bfm that balance precise control with temporally extended behavioral structure, together with
    \ac{sf}-guided interaction-preserving behaviors reuse.
\end{itemize}

\section{Related Work}
\label{sec:related-work}

\subsection{Humanoid Controller and Behavioral Foundation Models}

Humanoid control has increasingly moved from skill-specific policies toward shared controllers that support broad motor repertoires through a common conditioning interface.
Motion-conditioned approaches unify locomotion and whole-body behaviors through dense or sparse motion specifications
\cite{tessler2024maskedmimic,liao2025beyondmimic},
while recent systems extend unified control toward language-conditioned behaviors and humanoid loco-manipulation
\cite{jiang2025wholebodyvla,wei2026psi0}.
These works demonstrate that diverse behaviors can be consolidated within a single controller, but the supported behaviors are typically acquired through the controller's predefined training and conditioning interface.
Our work instead considers a complementary setting: integrating interaction behaviors acquired independently of this interface into an already-trained controller.

\ac{bfm}s expose a learned behavioral latent space as such a shared control interface.
BFM-Zero~\cite{li2025bfmzero} uses \ac{fb} to learn a promptable humanoid behavioral space, while subsequent behavioral inference methods show that expert behaviors can be recovered by optimizing within a frozen \ac{fb} latent space~\cite{pirotta2024fast}.
These results establish latent inference as an effective mechanism for behavior reproduction.
We build on this capability but study its use as an interface for interaction representation, where the latent sequence must capture temporally varying local control responses rather than only recover a stationary behavior or desired state trajectory.
This setting further exposes a trade-off in conventional \ac{fb} learning: the successor horizon favorable for precise local tracking differs from that required to preserve long-term behavior.
Our adaptive-horizon \ac{bfm} is designed to retain both properties within the same behavioral space.

\subsection{Humanoid Loco-Manipulation and Representation}

Humanoid loco-manipulation requires specifying not only the desired motion of the robot, but also how it should respond to physical interaction with objects and the environment.
A broad family of methods makes interaction explicit through structured task representations, including contact sequences, end-effector constraints, task-motion objectives, distance fields, and interaction-preserving references
\cite{zhang2025wococo,li2025hold,ren2025humanoid,lin2026lessmimic,yang2025omniretarget,chen2026scenebot,yu2026omnicontact}.
Such representations provide effective and often interpretable interfaces for learning diverse skills.
However, they specify interaction through variables chosen in advance; the resulting task description does not directly encode the state-dependent control response induced by contact and environment dynamics.

Alternatively, interaction responses can be learned directly through task-specific \ac{rl} or demonstrations
\cite{wang2023physhoi,wang2025skillmimic,yu2025skillmimic,zhang2025falcon,du2026cola,wang2026humanx,li2026haic,wang2026counterfactual}.
Teleoperation and human-video-based pipelines further broaden the sources from which such behaviors can be acquired
\cite{he2024omnih2o,ze2025twist,li2025clone,weng2025hdmi,nai2026humi,shi2026egohumanoid,wu2026sugar}.
These approaches can capture interaction-specific control responses through \ac{rl}, but the acquired behavior generally remains tied to the specialist policy or downstream training process.
In contrast, we decouple interaction acquisition from execution: demonstrated state-action responses are projected into the native latent interface of a frozen behavioral foundation controller, allowing independently acquired interactions to share a common executable substrate.

\section{Methodology}
\label{sec:method}

\begin{figure*}[t]
    \centering
    \includegraphics[width=\linewidth]{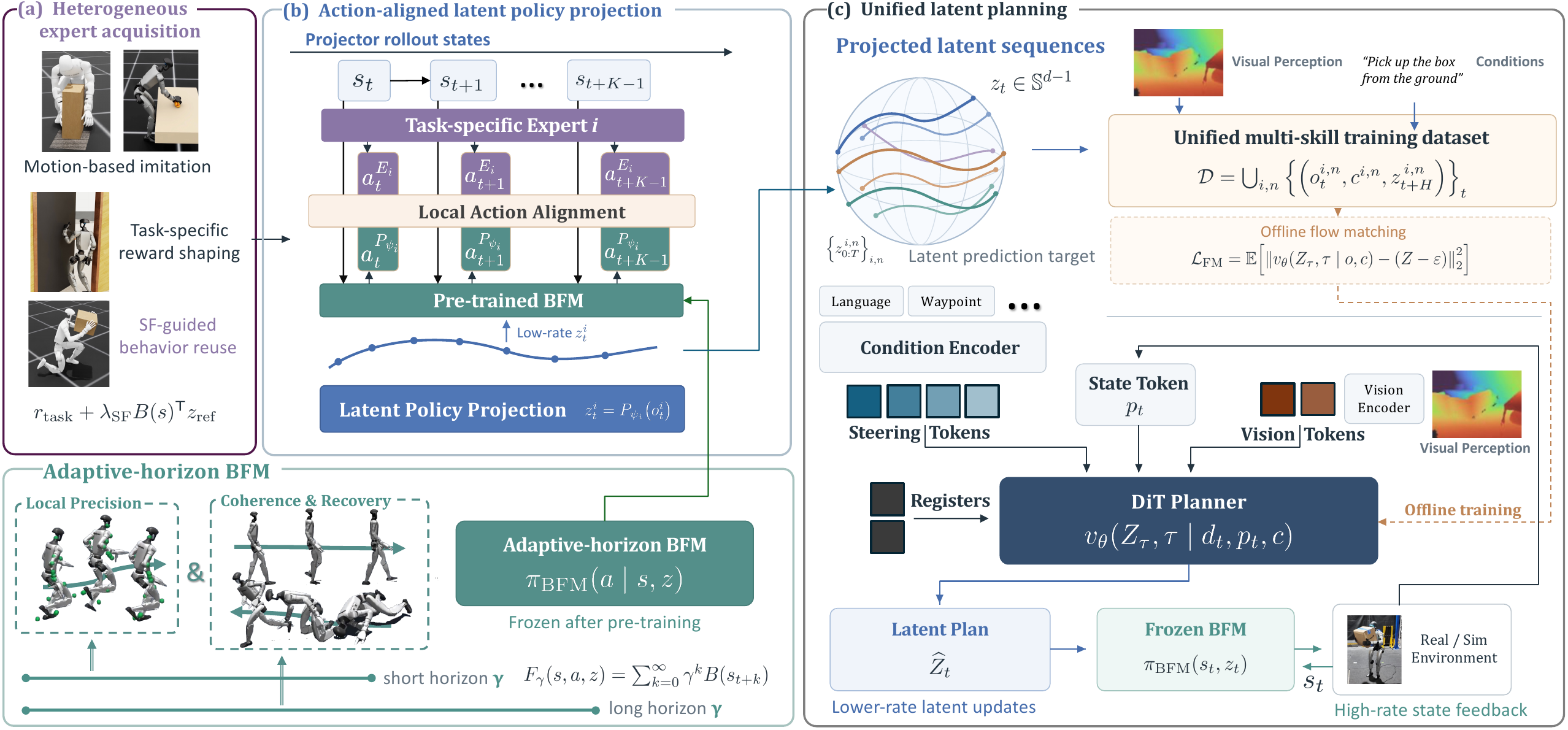}
    \caption{
    \textbf{Overview of \method.}
    \textbf{(a)} Heterogeneous interaction experts are acquired through motion imitation, reward shaping, or successor-feature-guided behavior reuse.
    \textbf{(b)} Action-aligned latent policy projection maps their short-horizon control responses into a frozen adaptive-horizon \ac{bfm}; closed-loop rollouts produce executable latent trajectories.
    \textbf{(c)} These trajectories train a unified perceptive planner entirely offline. Language commands or waypoint conditions are encoded as steering tokens, and a DiT planner predicts future latent trajectories. During deployment, the planner replans from onboard observations while the frozen \ac{bfm} provides high-frequency feedback.
    }
    \label{fig:pipeline}
\end{figure*}

\subsection{Overview}

Our goal is to express independently acquired interaction skills in a shared  behavioral interface that preserves their state-dependent physical realization and supports unified perceptive planning.
\method uses the latent space of a frozen \ac{bfm} as this interface and maps each interaction expert into it by aligning the expert's local control responses with the actions induced by \ac{bfm} latent commands.

As illustrated in~\cref{fig:pipeline}, \method proceeds in three stages. First, interaction skills are acquired independently with task-appropriate supervision.
Second, \ac{lpp} maps each expert into the frozen \ac{bfm} latent space by matching the actions induced by latent commands to the expert's local control responses.
The resulting projected policies are rolled out directly through the shared behavioral interface, after which the original experts are no longer required.
Finally, their closed-loop latent rollouts are aggregated to train a single perceptive flow planner entirely offline, conditioned on onboard observations and high-level steering commands.

This latent space must support both precise local interaction and temporally
extended whole-body behavior.
We therefore introduce \bfm, an adaptive-horizon \ac{bfm} trained with
multi-$\gamma$ \ac{fb} learning.
We consider
\begin{equation}
    \Gamma = \{\gamma_s,\gamma_l\}, \qquad \gamma_s < \gamma_l,
\end{equation}
where $\gamma_s$ emphasizes local transition fidelity and $\gamma_l$ preserves
longer-horizon behavioral structure.
We condition the forward representation $F$ on $\gamma$ while sharing the
backward representation $B$, and normalize the corresponding values as
\begin{equation}
    \widetilde{Q}_{\gamma}^{\mathrm{FB}}
    = (1-\gamma)Q_{\gamma}^{\mathrm{FB}}.
\end{equation}

The shared actor adaptively aggregates the horizon-specific objectives,
\begin{equation}
    J_{\pi}
    =
    \mathbb{E}
    \left[
        \tau \log
        \sum_{\gamma\in\Gamma}
        p(\gamma)
        \exp
        \left(
            \frac{
                \widetilde{Q}_{\gamma}^{\mathrm{FB}}
                (s,\pibfm(s,z),z)
            }{\tau}
        \right)
    \right],
\end{equation}
with gradient weights
\begin{equation}
    w_{\gamma}
    \propto
    p(\gamma)
    \exp\left(
        \frac{
            \widetilde{Q}_{\gamma}^{\mathrm{FB}}
            (s,\pibfm(s,z),z)
        }{\tau}
    \right).
\end{equation}
Here, $p(\gamma)$ defines a prior over horizons and $\tau$ controls the
aggregation sharpness.
Because the weights depend on the current state and latent command, the shared
actor can combine short- and long-horizon supervision within the same
behavioral space.

\subsection{Heterogeneous Skill Acquisition}
\label{sec:skill_acquisition}

\method allows interaction skills to be acquired with task-appropriate supervision,
observations, and training objectives.
We consider three complementary acquisition routes.

\paragraph{Motion-based acquisition}
For interactions with suitable humanoid--object motion references, such as grasping~\cite{xie2026grail}
and object pickup~\cite{weng2025hdmi}, we train task-specific experts by combining motion imitation with
interaction objectives.

\paragraph{Reward-shaping-based acquisition}
Some interactions admit a natural decomposition into successive stages and can
therefore be acquired directly through staged reward shaping.
For example, opening an articulated door~\cite{xue2026doorman} can be trained by progressively rewarding
approach, contact establishment, handle manipulation, and door opening.
This provides a direct route to task-specific experts without requiring motion
references, while allowing each stage to use task-appropriate observations and rewards.

\paragraph{\ac{sf}-guided behavior reuse}
The pretrained \ac{bfm} provides an additional route for extending acquired
interactions with whole-body capabilities already represented in its latent space.
For an interaction $I$, such as carrying box, we learn a latent adapter $A_{\psi_I}$ while keeping the \ac{bfm} frozen:
\begin{equation}
    \hat z_t = A_{\psi_I}(s_t,z^{N}),
    \qquad
    a_t = \pibfm(s_t,\hat z_t),
\end{equation}
where $z^{N}$ specifies the native behavior to be incorporated.

We optimize the adapter using the interaction reward together with a
successor-feature-guided preservation term,
\begin{equation}
    r_t
    =
    r_t^{I}
    +
    \lambda_{\mathrm{SF}} B(s_t)^\top z^{N}.
\end{equation}
For a fixed $z^{N}$,
\begin{equation}
\mathbb{E}
\left[
    \sum_{t=0}^{\infty}
    \gamma^t B(s_t)^\top z^{N}
\right]
=
\mathbb{E}
\left[
    \sum_{t=0}^{\infty}
    \gamma^t B(s_t)
\right]^\top z^{N},
\end{equation}
so the second term encourages the adapted rollout to preserve the behavior specified by $z^{N}$ while $r_t^{I}$ maintains the interaction.
This enables acquired interactions to be extended with native skills, such as running or kneeling while carrying.

Motion imitation, task-specific reward learning, and successor-feature-guided
behavior reuse each produce task-specific experts, all of which are
subsequently projected into the shared \ac{bfm} latent space through \ac{lpp}.

\begin{table*}[th!]
\vspace{1.5mm}
\centering
\caption{
Simulation success rates (\%) across heterogeneous loco-manipulation tasks.
``w/ appr.'' includes the approaching phase, whereas ``w/o appr.''
initializes the humanoid near the interaction state.
$^\dagger$ denotes a task-specific policy, and $^*$ denotes access to
privileged object states instead of egocentric observations. Mean is computed using the ``w/ appr.'' results.
}
\label{tab:sim_main}

\begin{tabular*}{\textwidth}{
@{\hspace{4pt}}
l
@{\extracolsep{\fill}}
cccccc
@{\hspace{4pt}}
}
\toprule
\multirow{2}{*}{Method}
& \multicolumn{2}{c}{Pickup Box}
& \multicolumn{2}{c}{Open Door}
& \multirow{2}{*}{Grasp Object}
& \multirow{2}{*}{Mean} \\
\cmidrule(lr){2-3}
\cmidrule(lr){4-5}
& w/ appr. & w/o appr.
& w/ appr. & w/o appr.
& & \\
\midrule

\rowcolor{gray!15}
\multicolumn{7}{l}{\textbf{Task-specific policies}} \\
Expert Policy$^\dagger$
& 100.0{\scriptsize$\pm$0.0} & 90.8{\scriptsize$\pm$0.5} & 89.3{\scriptsize$\pm$0.2} & 97.9{\scriptsize$\pm$0.8} & 97.0{\scriptsize$\pm$0.8} & 95.4 \\
\ac{lpp} w/ SONIC~\cite{luo2025sonic}$^\dagger$
& 0.0{\scriptsize$\pm$0.0} & 0.0{\scriptsize$\pm$0.0} & 83.1{\scriptsize$\pm$0.7} & 88.1{\scriptsize$\pm$0.6} & 5.1{\scriptsize$\pm$0.7} & 29.4 \\
LPP w/ Fixed-$\gamma$ BFM$^\dagger$
& 75.4{\scriptsize$\pm$0.2} & 80.2{\scriptsize$\pm$1.3}
& 61.8{\scriptsize$\pm$1.9} & 72.4{\scriptsize$\pm$0.7}
& 69.3{\scriptsize$\pm$1.2} & 68.8 \\
\textbf{\ac{lpp} w/ \bfm}$^\dagger$
& 81.9{\scriptsize$\pm$0.9} & 87.4{\scriptsize$\pm$2.4} & 83.5{\scriptsize$\pm$1.8} & 97.3{\scriptsize$\pm$0.4} & 52.9{\scriptsize$\pm$2.1} & 72.8 \\

\midrule
\rowcolor{gray!15}
\multicolumn{7}{l}{\textbf{Reference-prompted execution}} \\
SONIC~\cite{luo2025sonic}
& 0.0{\scriptsize$\pm$0.0} & 0.0{\scriptsize$\pm$0.0} & 0.4{\scriptsize$\pm$0.0} & 3.1{\scriptsize$\pm$0.0} & \underline{12.1}{\scriptsize$\pm$0.0} & 4.2 \\
FALCON~\cite{zhang2025falcon}
& 0.0{\scriptsize$\pm$0.0} & 0.0{\scriptsize$\pm$0.0} & 0.4{\scriptsize$\pm$0.0} & 2.7{\scriptsize$\pm$0.0} & 11.9{\scriptsize$\pm$0.3} & 4.1 \\
$B(s)$ Tracking
& 0.0{\scriptsize$\pm$0.0} & 0.0{\scriptsize$\pm$0.0} & 0.4{\scriptsize$\pm$0.0} & \underline{5.6}{\scriptsize$\pm$0.5} & \textbf{13.9}{\scriptsize$\pm$0.5} & \underline{4.8} \\
\textbf{\ac{lpp} Replay}
& \textbf{11.2}{\scriptsize$\pm$0.8} & \textbf{17.7}{\scriptsize$\pm$1.2} & \textbf{12.2}{\scriptsize$\pm$0.9} & \textbf{49.5}{\scriptsize$\pm$0.6} & 11.3{\scriptsize$\pm$1.1} & \textbf{11.6} \\

\midrule
\rowcolor{gray!15}
\multicolumn{7}{l}{\textbf{Perceptive planning}} \\
OmniContact$^*$~\cite{yu2026omnicontact}
& 29.9{\scriptsize$\pm$0.2} & \textbf{49.6}{\scriptsize$\pm$0.0} & -- & -- & -- & -- \\
LessMimic~\cite{lin2026lessmimic}
& \underline{38.3}{\scriptsize$\pm$0.0} & 28.5{\scriptsize$\pm$0.0} & -- & -- & -- & -- \\
Action Planner & 0.0{\scriptsize$\pm$0.0} & 0.0{\scriptsize$\pm$0.0}
& 0.0{\scriptsize$\pm$0.0} & 0.0{\scriptsize$\pm$0.0}
& 0.0{\scriptsize$\pm$0.0} & 0.0 \\
Kinematic Planner
& 0.0{\scriptsize$\pm$0.0} & 0.0{\scriptsize$\pm$0.0} & 0.0{\scriptsize$\pm$0.0} & 0.0{\scriptsize$\pm$0.0} & \underline{10.6}{\scriptsize$\pm$0.5} & \underline{3.5} \\
\textbf{Ours -- \method}
& \textbf{42.1}{\scriptsize$\pm$0.6} & \underline{44.1}{\scriptsize$\pm$1.0} & \textbf{53.5}{\scriptsize$\pm$4.8} & \textbf{51.2}{\scriptsize$\pm$1.0} & \textbf{30.3}{\scriptsize$\pm$2.3} & \textbf{42.0} \\

\bottomrule
\end{tabular*}
\end{table*}

\subsection{Action-Aligned Latent Policy Projection}
\label{sec:lpp}

Externally acquired experts must be converted into a representation suitable for
offline planner training.
Kinematic trajectories provide compact behavioral targets, but do not uniquely
determine the control responses required to realize an interaction.
Expert actions provide this missing physical information, yet directly modeling
high-frequency actions requires covering a broad state-dependent control
distribution.
We instead use the frozen \ac{bfm} to map expert behavior into a lower-rate
latent representation that preserves both its executable response and native
behavioral semantics.

For expert $\pi_i^{E}$, we learn a local projector $P_{\psi_i}$ that
maps the expert observation at update time $t_k$ to a latent,
\begin{equation}
    z_k = P_{\psi_i}(o_{t_k}).
\end{equation}
Between consecutive updates, we interpolate from the previous latent
$z_{k-1}$ to $z_k$ using SLERP to obtain a latent command at each control step,
\begin{equation}
    z_{t_k+h}
    =
    \operatorname{SLERP}\!\left(
        z_{k-1}, z_k, \alpha_h
    \right),
    \quad h=0,\ldots,H_P-1.
\end{equation}

We identify the latent command primarily through its induced actions.
Given expert actions
\begin{equation}
    a_t^{E_i}
    =
    \pi_i^{E}(o_{i,t}^{E}),
\end{equation}
we define the short-horizon action-alignment objective
\begin{equation}
\mathcal{L}_{\mathrm{act}}
=
\mathbb{E}_{P_{\psi_i}}
\left[
    \frac{1}{H_P}
    \sum_{h=0}^{H_P-1}
    \left\|
        \pibfm(s_{t_k+h},z_{t_k+h})
        -
        a_{t_k+h}^{E_i}
    \right\|_2^2
\right].
\end{equation}
Here, $P_{\psi_i}$ denotes the on-policy state distribution induced by the current projector executed through the frozen BFM. This directly constrains the projected latent by matching the expert's responses, rather than by a predefined kinematic or contact representation target.

Action alignment alone, however, can leave the latent underdetermined: different
commands may induce similar local actions while corresponding to different
behavioral directions in the \ac{bfm} space.
Recall that the \ac{bfm} latent $z$ acts as a policy index through the
forward--backward objective $F(s,a)^\top z$, while $B(s)$ provides the
corresponding state-side behavioral representation.
We therefore anchor each projected command to the native semantic representation
of its local state,
\begin{equation}
    \mathcal{L}_{\mathrm{sem}}
    =
    \mathbb{E}
    \left[
        \left\|
            z_k - B(s_{t_k})
        \right\|_2^2
    \right].
\end{equation}
The complete projection objective is
\begin{equation}
    \mathcal{L}_{\mathrm{proj}}
    =
    \mathcal{L}_{\mathrm{act}}
    +
    \lambda_{\mathrm{sem}}
    \mathcal{L}_{\mathrm{sem}}.
\end{equation}
The action term determines the latent according to the expert's required control
response, while the semantic term selects solutions consistent with the native
behavioral geometry of the \ac{bfm}.
This reduces action-equivalent latent ambiguity and yields more coherent latent
trajectories for downstream planning. To obtain a robust projection beyond nominal expert rollouts, we train each
$P_{\psi_i}$ independently using single-skill on-policy data aggregation~\cite{ross2011dagger}.

After projection, the original expert is no longer required.
We roll out $P_{\psi_i}$ under randomized initial states and environment
configurations and collect
\begin{equation}
    \mathcal{D}_{i}^{Z}
    =
    \left\{
        (d_t,p_t,u_t,z_t)
    \right\},
\end{equation}
where $d_t$ and $p_t$ denote the perceptive and proprioceptive observations
available to the downstream planner, and $u_t$ denotes the corresponding
high-level task condition.
These latent rollouts provide executable and semantically regularized targets
that can be combined across independently acquired skills for offline latent
planning.

\begin{figure*}[ht!]
    \centering
    \includegraphics[width=\linewidth]{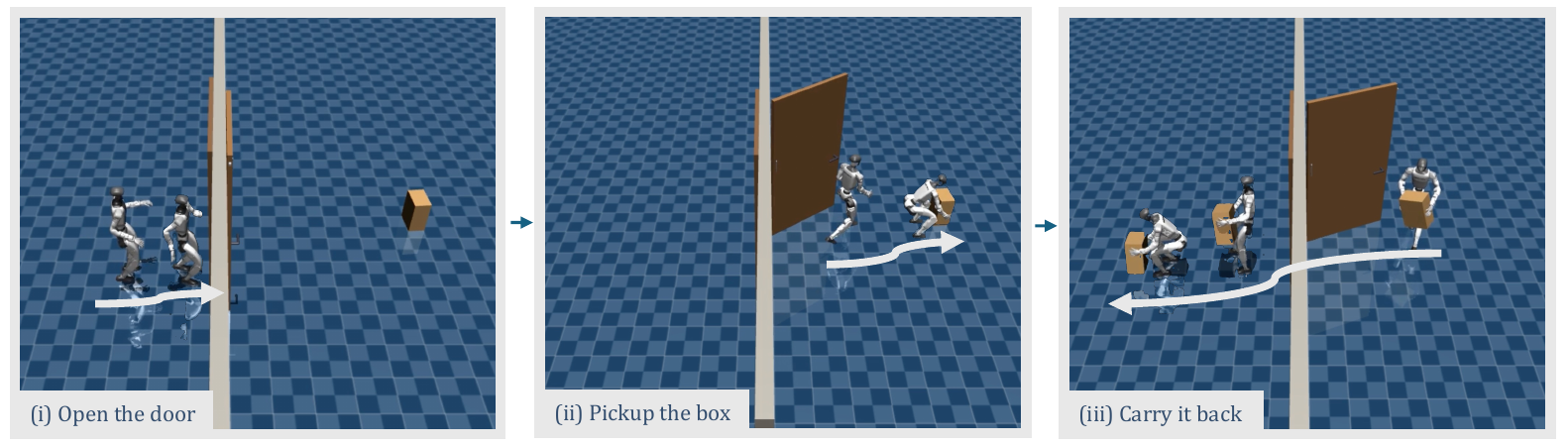}
    % \vspace{-20pt}
    \caption{
    \textbf{Skill chaining with \method.}
    In a single rollout, the policy sequentially opens the door in (i),
    picks up the box in (ii), and carries it back through the doorway in (iii),
    demonstrating closed-loop chaining of multiple interaction skills under changing
    high-level commands.
    }
    \label{fig:skill_chaining}
    % \vspace{-15pt}
\end{figure*}

\subsection{Unified Latent Flow Planning}
\label{sec:planner}

Given the collected latent rollouts, we train a single perceptive planner to
generate future behavioral commands entirely offline.
At each planning step, the policy is conditioned on onboard depth $d_t$,
proprioception $p_t$, and a compact steering token $c_t$:
\begin{equation}
    c_t
    =
    E_{\mathrm{steer}}(u_t).
\end{equation}
The steering input $u_t$ specifies the desired high-level behavior and can be
instantiated by different conditioning modalities, such as a language command or
a navigation waypoint.
This separates the representation of task intent from the perceptive latent
planner itself.

The planner predicts a horizon of future \ac{bfm} commands
\begin{equation}
    x_1
    =
    \operatorname{vec}
    \left(
        z_t,
        z_{t+1},
        \ldots,
        z_{t+H-1}
    \right)
\end{equation}
using conditional flow matching.
Given Gaussian noise $x_0 \sim \mathcal{N}(0,I)$
and flow time $\tau\sim\mathcal{U}(0,1)$, we construct the linear interpolation
$
    x_{\tau}
    =
    (1-\tau)x_0 + \tau x_1.
$
Then a conditional network predicts the clean latent, and is trained with
\begin{equation}
    \mathcal{L}_{\mathrm{flow}}
    =
    \mathbb{E}
    \left[
        \left\|
            \hat x_1-x_1
        \right\|_2^2
    \right].
\end{equation}
% The corresponding flow direction is obtained from the predicted endpoint as
% \begin{equation}
%     \hat v_{\eta}
%     =
%     \frac{
%         \hat x_1-x_{\tau}
%     }{
%         1-\tau
%     }.
% \end{equation}

At deployment, the planner repeatedly observes the current depth and
proprioceptive state, constructs the steering token from the desired high-level
condition, and generates a future latent trajectory.
The leading latent commands are executed through the frozen \ac{bfm}, after which
the planner replans from the newly observed state.
This receding-horizon execution allows the same policy to continuously adapt its
whole-body behavior to the perceived scene while following different steering
conditions.

Through this pipeline, skills originating from heterogeneous acquisition
procedures are converted into a common executable latent representation and
subsequently learned by a single offline perceptive planner.

\section{Experiments}

We evaluate \method through simulation and real-world experiments to answer four questions:

\textbf{Q1:} How faithfully can heterogeneous interaction experts be projected into the shared latent space of a frozen BFM?

\textbf{Q2:} Can these projected skills be jointly learned by a single planner for versatile humanoid loco-manipulation?

\textbf{Q3:} Can the unified planner chain projected skills and flexibly steer their execution through various conditions?

\textbf{Q4:} Which components of \method are critical for reliable latent projection and downstream planning?

For each projected skill, we collect 20K episodes, and merge them across tasks for \method training.
Language conditions are assigned according to the underlying task, while waypoint conditions are labeled from the robot trajectories in the collected rollouts.

\begin{figure*}[ht!]
    \centering
    \includegraphics[width=\linewidth]{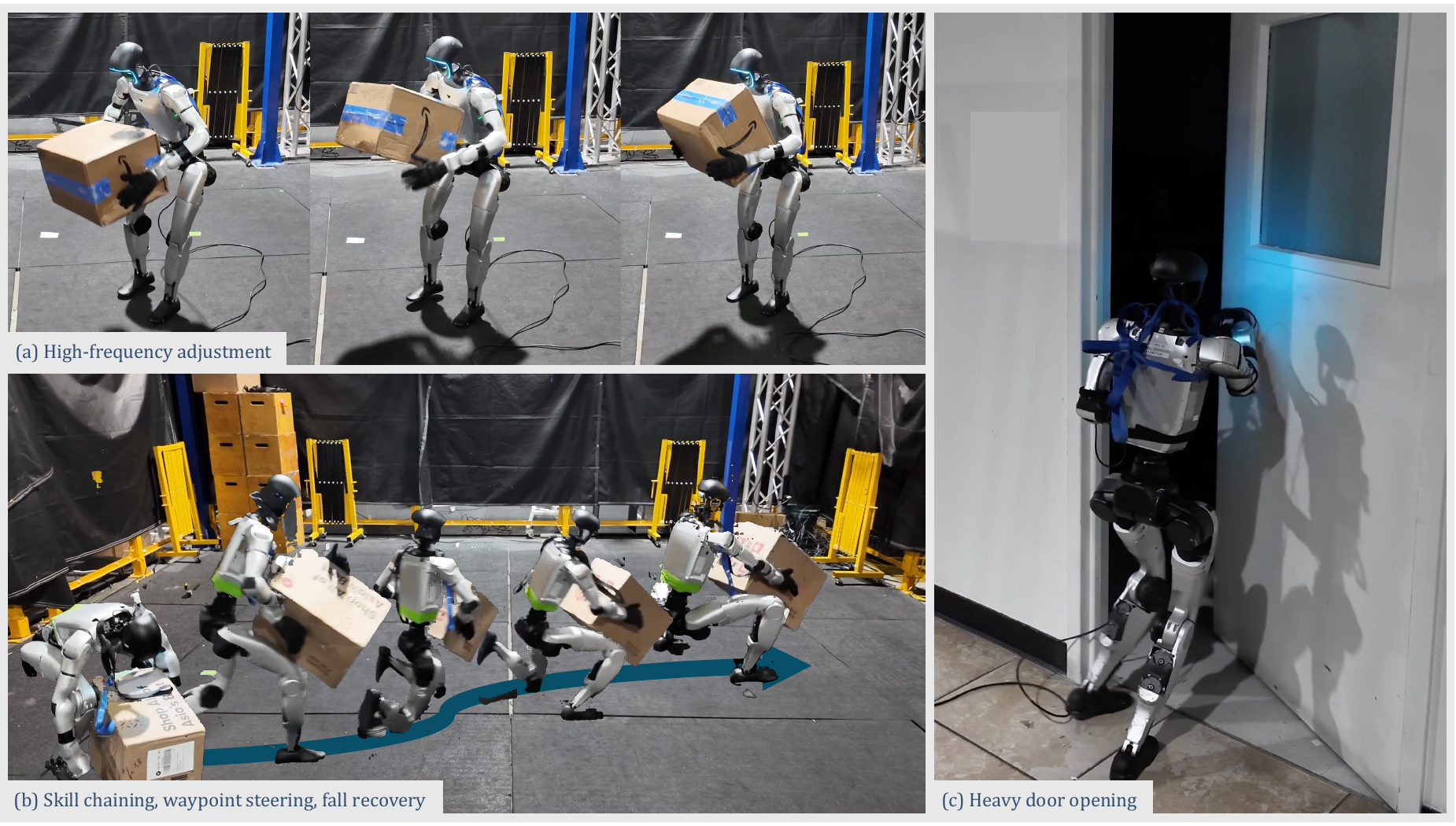}
    % \vspace{-20pt}
    \caption{
    \textbf{Qualitative real-world behaviors of \method.}
    (a) During box carrying, the robot reacts to an unstable grasp by briefly lifting the box and repositioning its hands underneath for more secure support, illustrating high-frequency closed-loop adjustment.
    (b) After picking up the box, the same policy follows waypoint commands while carrying it; after stumbling, the robot stabilizes on its knees and recovers before continuing, demonstrating interaction-preserving skill chaining, steering, and recovery.
    (c) For a heavy door, the robot leans its whole body into the interaction to generate sufficient pushing force, showing adaptive whole-body realization of contact-rich tasks.
    }
    \label{fig:qualitative}
    % \vspace{-15pt}
\end{figure*}

\subsection{Simulation Evaluation}

We evaluate in IsaacSim, with \textbf{128 variations per task}: box size and
initial placement for pickup, door geometry for opening, and object identity for
grasping. 
We do not report on-policy multi-expert distillation because the experts use
different environments, observation spaces, and codebases, and therefore do
not share the online rollout interface required for joint distillation.

\paragraph{Action-aligned latent projection}
To answer \textbf{Q1}, we compare each task-specific expert with its \ac{lpp}
projector in~\cref{tab:sim_main}.
\ac{lpp} largely preserves expert performance on box pickup and door opening,
while fine-grained grasping is more challenging, dropping from $97.0\%$ to
$52.9\%$.

We further compare representations using targets derived from the same successful
expert trajectories.
Reference-tracking baselines, including the \ac{bfm}'s own backward
representation $B(s)$, perform poorly, whereas action-aligned projected replay
achieves substantially higher success on box pickup and door opening.
Notably, FALCON~\cite{zhang2025falcon}, despite learning force-adaptive
end-effector tracking under randomized external forces, achieves only
$0.4\%/2.7\%$ on door opening.
These results show that kinematic, behavioral, or generic force-adaptive
tracking alone does not recover the task-specific control responses required for
interaction, motivating action-aligned latent projection.

\paragraph{Unified perceptive planning}
To answer \textbf{Q2}, we compare replaying \ac{lpp}-produced latent trajectories
with the unified perceptive planner.
\method improves mean success from $11.6\%$ to $42.0\%$, showing that
high-frequency \ac{bfm} feedback alone is insufficient and that task-level
latent replanning remains important.
Under the same perceptive planning setup, the Action Planner fails on all tasks,
while the Kinematic Planner reaches only $3.5\%$ mean success.
Directly planning low-level actions removes high-frequency state feedback
between planner updates, whereas predicting kinematic targets does not specify the
interaction-specific control response, which leads to failure on contact-rich tasks; \method retains both through latent planning over \ac{bfm}.

\paragraph{Skill chaining and conditional steering}
To answer \textbf{Q3}, we further test whether the unified planner can compose
projected skills under changing high-level commands.
As shown in~\cref{fig:skill_chaining}, the same policy sequentially opens the
door, picks up the box, and carries it back through the doorway within a single
rollout, demonstrating closed-loop chaining of heterogeneous interaction skills.

% \paragraph{Skill Chaining and Conditional Steering}

% \yutang{Finalize the evaluation protocol for Q3, including skill chaining,
% language-conditioned switching, and waypoint steering.}

\subsection{Real-World Evaluation}

We further deploy \method on a Unitree G1 equipped with an onboard RealSense
depth camera to evaluate real-world transfer of the unified planner
(\textbf{Q2}) and interaction chaining and steering (\textbf{Q3}).
Both the planner and \bfm run on a single NVIDIA RTX PRO 5000 Blackwell laptop.
With TensorRT, the planner runs at $5$\,Hz using 10 Heun steps and \ac{rtc}
($\sim$20\,ms per inference), while \bfm runs at $50$\,Hz
($\sim$1\,ms per forward pass).

\cref{tab:real_main} reports success rates for box pickup, door opening, and
object grasping. All experts, latent projectors, and the unified planner are
trained entirely in simulation.
\method succeeds in $6/10$ box-pickup trials and $4/10$ door-opening trials,
demonstrating direct sim-to-real transfer of the latent planning interface.
Object grasping is more challenging ($2/10$), where inaccurate depth estimates
for the small object compound the fine-grained control difficulty.
Beyond task success, \cref{fig:qualitative} highlights the key advantage of our
representation: planned latents remain grounded in closed-loop physical
responses.
When the box becomes unstable, the robot repositions its hands underneath for
support in (a); when opening a heavy door, it leans its whole body into the
interaction to generate sufficient force in (c).

To answer \textbf{Q3}, we further evaluate post-pickup behavior chaining and
waypoint steering.
After picking up a box, the same planner follows root-waypoint commands while
maintaining the grasp; when disturbed during carrying, it can stabilize on its
knees, recover, and continue execution, as shown in
\cref{fig:qualitative} (b).
These results show that the unified
planner can chain interaction and locomotion behaviors while remaining
steerable through high-level waypoint commands.

\begin{table}[t!]
\centering
\vspace{1.5mm}
\caption{
Real-world success rates of the unified \method planner on the Unitree G1.
}
\label{tab:real_main}

\begin{tabular}{@{}lcccc@{}}
\toprule
Method & Pickup Box & Open Door & Grasp Object & Mean \\
\midrule
\method & 6/10 & 4/10 & 2/10 & 40\% \\
\bottomrule
\end{tabular}
\end{table}

\begin{table}[t!]
    \centering
    \caption{Ablation of multi-horizon BFM tracking and behavior embedding performance on 100STYLE~\cite{mason2022local}.}
    \label{tab:multihorizon}
    \begin{tabular*}{\columnwidth}{
    @{\hspace{\tabcolsep}}
    l
    @{\extracolsep{\fill}}
    ccccc
    @{\hspace{\tabcolsep}}
    }
    \toprule
    \multirow{2}{*}{Method} &
    \multirow{2}{*}{Joint-MAE $\downarrow$} &
    \multicolumn{4}{c}{BE-EMD $\downarrow$} \\
    \cmidrule(lr){3-6}
     & & 0.2\,s & 0.5\,s & 1\,s & 2\,s \\
    \midrule
    SONIC~\cite{luo2025sonic}    & 0.1315 & -- & -- & -- & -- \\
    BFM $\gamma_{0.8}$  & 0.1107 & \textbf{0.038} & 0.200 & 0.495 & 0.792 \\
    BFM $\gamma_{0.98}$ & 0.1327 & 0.048 & 0.213 & 0.451 & 0.601 \\
    \bfm                         & \textbf{0.1020} & 0.049 & \textbf{0.195} & \textbf{0.422} & \textbf{0.590} \\
    \bottomrule
    \vspace{-9mm}
    \end{tabular*}
\end{table}

\subsection{Ablation and Analysis}

To answer \textbf{Q4}, we analyze the two components underlying our latent
planning interface: the adaptive-horizon BFM and the action-aligned latent
representation.

\paragraph{Effect of the adaptive-horizon BFM}
We first evaluate how the BFM temporal horizon affects both behavior representation and downstream projection.

For behavior representation, we evaluate \bfm using two complementary metrics.
Joint-MAE measures instantaneous motion-tracking accuracy, while BE-EMD measures how well a fixed latent command preserves a reference behavior over increasing embedding temporal horizons: for each reference clip, we infer a single latent through \ac{lpp}, roll out the BFM under this fixed command, and measure its discrepancy from the reference.

As shown in~\cref{tab:multihorizon}, the short-horizon model $\gamma=0.8$ achieves the lowest BE-EMD at $0.2$\,s, whereas $\gamma=0.98$ better preserves behavior over longer horizons.
\bfm achieves the lowest Joint-MAE and the lowest BE-EMD at $0.5$, $1$, and $2$\,s.
Adaptive multi-$\gamma$ learning therefore provides a better trade-off between precise local control and temporally extended behavioral structure than either fixed horizon alone.

This advantage also transfers to interaction projection:
in~\cref{tab:sim_main}, replacing the fixed-$\gamma$ BFM
($\gamma=0.98$) with \bfm increases mean \ac{lpp} success from
$68.8\%$ to $72.8\%$, with substantial gains on box pickup and door opening.

\paragraph{Effect of action-aligned latent representation}
The reference-prompted results in~\cref{tab:sim_main} isolate the importance of action alignment.
$B(s)$ tracking and projected replay are derived from the same successful expert trajectories, but the former identifies behavior from the expert state trajectory whereas \ac{lpp} identifies latent commands from the expert's action responses.
Their large performance difference on pickup box and door opening shows that behavioral-state matching alone does not retain interaction-critical feedback.
At the other extreme, the Action Planner fails in all tasks, showing that
low-rate action planning cannot replace the high-frequency state feedback
provided by the frozen \ac{bfm}.

The gap between the feedback-conditioned \ac{lpp} projector and fixed projected replay further shows that projected behaviors cannot generally be reduced to open-loop latent trajectories.
Conversely, the improvement from projected replay to \method shows that a unified perceptive planner can recover task-level latent feedback from onboard observations.
Together, these comparisons separate the roles of action-aligned representation, latent feedback, and perceptive replanning in \method.

% \begin{table}[t]
% \centering
% \caption{
% Evaluation of skill chaining.
% The unified policy maintains the grasped box after picks up, while executing different subsequent behaviors.
% }
% \label{tab:behavior_chaining}

% \begin{tabular*}{\columnwidth}{@{}l@{\extracolsep{\fill}}cc@{}}
% \toprule
% Method
% & Pickup $\rightarrow$ Kneel
% & Open Door $\rightarrow$ Pickup \\
% \midrule
% \method
% & -- & -- \\
% \bottomrule
% \end{tabular*}
% \end{table}

\section{Conclusions and Limitations}

We presented \method, an action-aligned latent planning framework for
versatile humanoid loco-manipulation. 
\ac{lpp} projects heterogeneous interaction experts into a frozen BFM by matching their short-horizon control responses, producing executable latent
trajectories for unified offline perceptive planning.
Adaptive multi-$\gamma$ \ac{fb} learning supports both precise control and
temporally extended behavior, while successor-feature-guided learning enables interaction-preserving behavior reuse.
Our simulation and real-world experiments demonstrate effective expert
projection and perception-conditioned execution across diverse interaction
tasks.

While our approach enables versatile humanoid loco-manipulation through
action-aligned latent planning, extending the underlying BFM toward more
dynamic whole-body behaviors, such as stair traversal, remains worth
exploring.
Moreover, incorporating richer visual and tactile feedback together with
online latent adaptation could support interaction under more diverse contact conditions and facilitate the continual integration of new skills.
We hope our work provides insights into scalable humanoid skill acquisition
and opens new directions for unified whole-body planning.

\FloatBarrier

{
\bibliographystyle{IEEEtran}
\bibliography{reference_header,reference}
}

\clearpage
\appendices
\section{Adaptive-Horizon Behavioral Foundation Model}
\label{app:bfm}

This appendix follows the order of the methodology in the main paper. We first
detail the adaptive-horizon BFM that defines the shared executable space, then
the three routes for skill acquisition, action-aligned latent policy projection,
and finally unified offline latent planning.

The same BFM latent space serves two roles in \method: it is the executable
interface used to project interaction experts and the prediction space of the
unified planner. Projection favors locally precise responses, while latent
planning benefits from commands with longer temporal coherence. The
adaptive-horizon BFM is designed to support both requirements within one shared
behavioral space.

\subsection{Executable Behavioral Interface and BFM Observations}
\label{app:control}

The body controller commands the 29 actuated joints of the Unitree G1 at
$50$\,Hz, with control interval $\Delta t=0.02$\,s. Tasks using Dex3 hands
additionally predict 14 hand-joint targets, seven per hand. These hand targets
are produced by a separate branch and are not part of the BFM body action.

Let $d_z$ denote the BFM latent dimension. Executable body commands lie on
\begin{equation}
    \mathcal Z
    =
    \{z\in\mathbb R^{d_z}:\|z\|_2=\sqrt{d_z}\},
    \quad
    \Pi_{\mathcal Z}(v)
    =
    \sqrt{d_z}\frac{v}{\|v\|_2},
    \label{eq:app_sphere}
\end{equation}
with a fixed nonzero reference vector used only for the degenerate zero-norm
case. We use $d_z=256$. The fixed-radius parameterization keeps the average
squared coordinate magnitude independent of the latent dimension.

The BFM actor consumes the current robot state together with causal history.
Its current proprioceptive state is
\begin{equation}
    s_t^{\mathrm{prop}}
    =
    \left[
    q_t-q^{\mathrm{default}};
    \dot q_t;
    g_t^{\mathrm{body}};
    0.25\,\omega_t^{\mathrm{body}}
    \right]
    \in\mathbb R^{64},
    \label{eq:app_bfm_state}
\end{equation}
where $g_t^{\mathrm{body}}$ is the gravity direction and
$\omega_t^{\mathrm{body}}$ is the body angular velocity. The actor also
receives the previous scaled body action and nine causal proprioceptive/action
history frames. Each history frame contains 93 entries: 29 action entries,
three angular velocities, 29 joint positions, 29 joint velocities, and three
gravity entries. The actor therefore receives $64+29+9\times93=930$
proprioceptive/action coordinates before the latent command is appended.

The backward representation additionally receives 463 dimensions of
heading-frame whole-body kinematics during BFM training and simulation-based
projection. These privileged quantities are not consumed by the deployed BFM
actor. Domain-randomization variables and external forces may likewise enter
training-time representations and critics without entering the deployed policy.

\subsection{Multi-Horizon Forward--Backward Representation}

For a latent-conditioned policy, the forward representation admits the
successor-feature interpretation
\begin{equation}
    F_\gamma(s_t,a_t,z)
    \approx
    \mathbb E
    \left[
        \sum_{h=0}^{\infty}
        \gamma^h B(s_{t+h+1})
        \;\middle|\;
        s_t,a_t,\pibfm(\cdot,z)
    \right].
    \label{eq:app_successor}
\end{equation}
The corresponding FB value is
$Q_\gamma^{\mathrm{FB}}(s,a,z)=F_\gamma(s,a,z)^\top z$.
We condition the forward representation on $\gamma$ while sharing both the
backward map $B$ and the executable actor across horizons. For every replay
transition, a horizon is sampled from the configured set $\Gamma$, and the same
$\gamma$ is used in the online prediction and bootstrapped target. The output
of $B$ is projected to $\mathcal Z$.

For a minibatch of $n$ transitions $(s_i,a_i,s_i^+,d_i)$, with termination
indicator $d_i$, define
\begin{align}
    M_{ij}
    &=F_{\gamma_i}(s_i,a_i,z_i)^\top B(s_j^+),\\
    R_{ij}
    &=M_{ij}-\gamma_i(1-d_i)\bar M_{ij},
\end{align}
where $\bar M_{ij}$ is a stopped-gradient target constructed from target
representations and a next action sampled from the latent-conditioned actor.
Suppressing ensemble notation, the FB minibatch objective is
\begin{equation}
    \mathcal L_{\mathrm{FB}}
    =
    \frac{\sum_{i\neq j}R_{ij}^2}{2n(n-1)}
    -
    \frac{1}{n}\sum_i R_{ii}.
    \label{eq:app_fb_batch}
\end{equation}
The shared backward map is regularized through
\begin{equation}
    \mathcal L_{\mathrm{orth}}
    =
    \frac{\sum_{i\neq j}(b_i^\top b_j)^2}{2n(n-1)}
    -
    \frac{1}{n}\sum_i\|b_i\|_2^2,
    \qquad b_i=B(s_i^+),
\end{equation}
and the representation update minimizes
$\mathcal L_{\mathrm{FB}}+\lambda_{\mathrm{orth}}\mathcal L_{\mathrm{orth}}$.
Motion-discriminator and auxiliary-critic objectives follow the underlying BFM
training procedure; the multi-horizon modification applies to the FB
representation and the shared actor objective below.

\subsection{Adaptive Actor Objective}

Values at different discounts have different natural scales. We therefore use
the normalized horizon-specific objective
\begin{equation}
    V_\gamma(s,z)
    =
    (1-\gamma)
    Q_\gamma^{\mathrm{FB}}
    \bigl(s,\pibfm(s,z),z\bigr).
    \label{eq:app_normalized_value}
\end{equation}
The factor $(1-\gamma)$ removes the leading discounted-sum scale, making the
short- and long-horizon terms directly comparable. The shared actor maximizes
\begin{equation}
    J_{\mathrm{FB}}(s,z)
    =
    \tau\log
    \sum_{\gamma\in\Gamma}
    p(\gamma)
    \exp\!\left(\frac{V_\gamma(s,z)}{\tau}\right),
    \label{eq:app_lse}
\end{equation}
where $p(\gamma)\geq0,\sum_\gamma p(\gamma)=1$, and $\tau>0$. Actor
gradient is
\begin{align}
    \nabla_\theta J_{\mathrm{FB}}
    &=
    \sum_{\gamma\in\Gamma}
    w_\gamma(s,z)\nabla_\theta V_\gamma(s,z),\\
    w_\gamma(s,z)
    &=
    \frac{p(\gamma)\exp(V_\gamma(s,z)/\tau)}
    {\sum_{\gamma'\in\Gamma}
      p(\gamma')\exp(V_{\gamma'}(s,z)/\tau)}.
    \label{eq:app_horizon_weights}
\end{align}
Thus the update remains a convex combination of the horizon-specific policy
gradients, but the combination is selected from the current state and latent
command rather than fixed globally.

\paragraph{Stability of the adaptive FB-to-actor objective}
The relevant stability property is not only that the horizon aggregator is
non-expansive, but that it does not amplify perturbations introduced through
the FB backup.
Let
\begin{equation}
    \widetilde Q_\gamma
    =
    (1-\gamma)Q_\gamma^{\mathrm{FB}},
\end{equation}
and collect the normalized action values as
$\widetilde{\mathbf Q}
=
[\widetilde Q_\gamma]_{\gamma\in\Gamma}$.
Holding the current backward map $B$ fixed, let $r_z^B$ denote the
corresponding latent-conditioned reward.
For a fixed latent policy $\pi$, the normalized FB Bellman
operator is
\begin{equation}
\begin{split}
    &(\mathcal B^\pi\widetilde{\mathbf Q})_\gamma(s,a,z)
    \\
    &\quad =
    \mathbb E\!\left[
        (1-\gamma)r_z^B(s')
        +
        \gamma(1-d)
        \widetilde Q_\gamma
        \bigl(s',\pi(s',z),z\bigr)
        \mid s,a
    \right].
\end{split}
\label{eq:app_fb_operator}
\end{equation}
Equip the product space with the norm
\begin{equation}
    \|\widetilde{\mathbf Q}\|_{\Gamma,\infty}
    =
    \max_{\gamma\in\Gamma}
    \|\widetilde Q_\gamma\|_\infty .
\end{equation}
For any two value collections $\widetilde{\mathbf Q}$ and
$\widetilde{\mathbf Q}'$, the reward terms cancel, yielding
\begin{equation}
\begin{split}
    \|\mathcal B^\pi\widetilde{\mathbf Q}
      -
      \mathcal B^\pi\widetilde{\mathbf Q}'\|_{\Gamma,\infty}
    \leq
    \gamma_{\max}
    \|\widetilde{\mathbf Q}
      -
      \widetilde{\mathbf Q}'\|_{\Gamma,\infty},
\end{split}
\label{eq:app_fb_contraction}
\end{equation}
where $\gamma_{\max}=\max\Gamma<1$.
Thus, for fixed $B$ and $\pi$, the multi-horizon FB Bellman backup is a
$\gamma_{\max}$-contraction.

We next define the adaptive actor-value functional
\begin{equation}
    (\mathcal A_\tau^\pi\widetilde{\mathbf Q})(s,z)
    =
    \tau\log
    \sum_{\gamma\in\Gamma}
    p(\gamma)
    \exp\!\left(
        \frac{
        \widetilde Q_\gamma(s,\pi(s,z),z)
        }{\tau}
    \right).
    \label{eq:app_actor_operator}
\end{equation}
The log-sum-exp map is monotone and translation equivariant.
Let
\begin{equation}
    \delta
    =
    \|\widetilde{\mathbf Q}
      -
      \widetilde{\mathbf Q}'\|_{\Gamma,\infty}.
\end{equation}
Then, for every $(s,a,z)$ and $\gamma\in\Gamma$,
\begin{equation}
    \widetilde Q'_\gamma(s,a,z)-\delta
    \leq
    \widetilde Q_\gamma(s,a,z)
    \leq
    \widetilde Q'_\gamma(s,a,z)+\delta .
\end{equation}
Monotonicity and translation equivariance therefore imply
\begin{equation}
    \|\mathcal A_\tau^\pi\widetilde{\mathbf Q}
      -
      \mathcal A_\tau^\pi\widetilde{\mathbf Q}'\|_\infty
    \leq
    \|\widetilde{\mathbf Q}
      -
      \widetilde{\mathbf Q}'\|_{\Gamma,\infty}.
    \label{eq:app_actor_nonexpansive}
\end{equation}
Hence the adaptive actor-value functional is non-expansive with respect to
perturbations of the horizon-specific FB values.

The same bound holds for the exact improved-value functional
\begin{equation}
    (\mathcal I_\tau\widetilde{\mathbf Q})(s,z)
    =
    \max_a
    \tau\log
    \sum_{\gamma\in\Gamma}
    p(\gamma)
    \exp\!\left(
        \frac{\widetilde Q_\gamma(s,a,z)}{\tau}
    \right),
    \label{eq:app_improved_value}
\end{equation}
since
\begin{equation}
    \left|
        \max_a f(a)-\max_a g(a)
    \right|
    \leq
    \max_a |f(a)-g(a)|.
\end{equation}
This statement concerns the improved value functional rather than the
$\arg\max$ policy map itself.

Combining~\eqref{eq:app_fb_contraction}
and~\eqref{eq:app_actor_nonexpansive} gives
\begin{equation}
\begin{split}
    \|\mathcal A_\tau^\pi
      \mathcal B^\pi\widetilde{\mathbf Q}
      -
      \mathcal A_\tau^\pi
      \mathcal B^\pi\widetilde{\mathbf Q}'\|_\infty
    \leq
    \gamma_{\max}
    \|\widetilde{\mathbf Q}
      -
      \widetilde{\mathbf Q}'\|_{\Gamma,\infty}.
\end{split}
\label{eq:app_fb_actor_contraction}
\end{equation}
Therefore, adaptive horizon aggregation preserves the contractive sensitivity
of the fixed-policy FB backup: perturbations in the horizon-specific FB values
are not amplified by the joint actor objective. Consequently, relative to a
single-horizon objective with $\gamma=\gamma_{\max}$, adaptive aggregation
introduces no worse worst-case sensitivity to critic perturbations while
allowing the shared policy to exploit value estimates at multiple temporal
horizons.

\paragraph{Adaptive aggregation versus a fixed weighted sum}
A natural multi-$\gamma$ alternative is the fixed convex average
\begin{equation}
    J_{\mathrm{WS}}(s,z)
    =
    \sum_{\gamma\in\Gamma}p(\gamma)V_\gamma(s,z).
    \label{eq:app_weighted_sum}
\end{equation}
Both $J_{\mathrm{WS}}$ and \eqref{eq:app_lse} are scale controlled after the
$(1-\gamma)$ normalization; non-expansiveness alone therefore does not distinguish
the two. The key difference is that the weighted-sum baseline uses the same
horizon weights for every state and behavior, whereas \eqref{eq:app_lse} adapts
the effective weights to the current horizon-specific values. As
$\tau\rightarrow0$, the update concentrates on the largest normalized value; for
large $\tau$, its gradient approaches the prior-weighted average. To isolate this
effect from simply exposing the actor to multiple discounts, we additionally
compare against \eqref{eq:app_weighted_sum} using the same multi-horizon
representation. The adaptive objective empirically outperforms this fixed weighted-sum
baseline, supporting state- and behavior-dependent horizon weighting rather than
a fixed global mixture.

For intuition, a discount $\gamma$ at control interval $\Delta t$ has effective
duration approximately $\Delta t/(1-\gamma)$. At $50$\,Hz, $\gamma=0.8$
corresponds to approximately $0.1$\,s and $\gamma=0.98$ to approximately
$1$\,s. This is an interpretation of the temporal weighting, not a hard
execution horizon for the latent command.

\subsection{Architecture and Optimization}

The actor, forward representations, and critics use residual MLP backbones.
The forward representations and critics use two ensemble members; the actor is
a single shared network conditioned on the current latent and outputs 29 body
actions. The backward encoder maps proprioceptive and privileged kinematic
features to the normalized behavioral space. The scalar discount is embedded
by a small MLP before entering the forward representation and is not an input to
the actor. Details are shown in~\cref{tab:app_bfm_arch}.

\begin{table}[t]
    \centering
    \caption{BFM architecture used in our implementation.}
    \label{tab:app_bfm_arch}
    \small
    \begin{tabular*}{\columnwidth}{
    @{\hspace{\tabcolsep}}
    l
    @{\extracolsep{\fill}}
    cc
    @{\hspace{\tabcolsep}}
    }
        \toprule
        Component & Hidden width & Hidden layers \\
        \midrule
        Body actor & 2048 & 6 \\
        Forward representation & 2048 & 6 \\
        Critics & 2048 & 6 \\
        Backward encoder & 1024 & 1 \\
        Motion discriminator & 1024 & 3 \\
        \midrule
        Executable latent dimension & \multicolumn{2}{c}{256} \\
        Body action dimension & \multicolumn{2}{c}{29} \\
        Forward/critic ensemble size & \multicolumn{2}{c}{2} \\
        \bottomrule
    \end{tabular*}
\end{table}

The BFM is frozen during expert projection, projected-data collection, and
offline planner training. These stages therefore use the same latent geometry,
state convention, and body-action decoder.

\begin{table}[t]
    \centering
    \caption{BFM optimization parameters. Target networks use Polyak updates
    $\bar\theta\leftarrow(1-\rho)\bar\theta+\rho\theta$.}
    \label{tab:app_bfm_train}
    \small
    \begin{tabular*}{\columnwidth}{
    @{\hspace{\tabcolsep}}
    l
    @{\extracolsep{\fill}}
    c
    @{\hspace{\tabcolsep}}
    }
        \toprule
        Parameter & Value \\
        \midrule
        Forward-map learning rate & $3\times10^{-4}$ \\
        Backward-map learning rate & $10^{-5}$ \\
        Actor / critic learning rate & $3\times10^{-4}$ \\
        Discriminator learning rate & $10^{-5}$ \\
        Minibatch size & 1024 \\
        Training updates & 100,000 \\
        Horizon set $\Gamma$ & $\{0.8,0.98\}$ \\
        Horizon prior $p(0.8)/p(0.98)$ & $0.625/0.375$ \\
        Aggregation temperature $\tau$ & 1 \\
        Replay capacity & $5.12\times10^6$ \\
        FB target update rate $\rho$ & 0.01 \\
        Critic target update rate $\rho$ & 0.005 \\
        Backward orthogonality coefficient & 100 \\
        \bottomrule
    \end{tabular*}
\end{table}

\section{Heterogeneous Skill Acquisition}
\label{app:acquisition}

As in the main paper, skill acquisition is deliberately decoupled from the
shared representation. Experts may use different supervision, observations,
and training procedures; projection only requires access to the resulting robot
actions. We use motion-based and reward-shaped experts for the base
interactions, and SF-guided behavior reuse when we specifically want to create
diverse compositions between an interaction and native BFM behaviors.

\subsection{Task-Specific Interaction Experts}

Motion-based experts combine humanoid motion tracking with object and contact
objectives. For box pickup and object grasping, the tracked quantities include
both body motion and manipulated-object pose, while contact rewards encourage
the demonstrated interaction to be realized physically. The door expert is
trained with stage-specific objectives~\cite{xue2026doorman} for approach, contact establishment,
handle manipulation, and opening. These experts may use different privileged
observations and rewards; none of those task-specific inputs are
required by the unified planner after projection.

For reproducibility, the HDMI-style motion expert uses a privileged observation
encoder with a 256-dimensional embedding and a reference-action residual actor.
Its actor hidden widths are $[512,256,256]$, critic hidden widths are
$[512,256,128]$, and the activation is Mish. The PPO settings for this route
are summarized in \cref{tab:app_ppo}.

\begin{table}[t]
    \centering
    \caption{PPO parameters for the motion expert and the SF-guided composition
    adapter.}
    \label{tab:app_ppo}
    \small
    \begin{tabular*}{\columnwidth}{
    @{\hspace{\tabcolsep}}
    l
    @{\extracolsep{\fill}}
    cc
    @{\hspace{\tabcolsep}}
    }
        \toprule
        Parameter & Motion expert & SF adapter \\
        \midrule
        Rollout steps / environment & 32 & 32 \\
        Learning rate & $3\!\times\!10^{-4}$ & $10^{-5}$ \\
        Optimization epochs & 3 & 5 \\
        Minibatches / rollout & 8 & 4 \\
        PPO clipping parameter & 0.2 & 0.2 \\
        Discount & 0.99 & 0.99 \\
        GAE parameter & 0.95 & 0.95 \\
        Entropy coefficient & 0.001 & 0 \\
        Gradient-norm limit & 1 & 1 \\
        \bottomrule
    \end{tabular*}
\end{table}

\subsection{SF-Guided Behavior Reuse for Skill Composition}
\label{app:sf_comp}

The third acquisition route targets interaction--behavior composition. Given
an acquired interaction, we use behaviors already represented by the BFM to
create additional physically valid variants: the interaction reward preserves
the task-specific physical requirements, while a native BFM latent specifies
the whole-body behavior to be expressed simultaneously. This expands one
interaction into a family of compositions without requiring a separate motion
reference or task-specific expert for every combination.

In our experiments, we instantiate this idea with box carrying. By varying the
requested native latent, the same carrying interaction can be composed with
crouching, kneeling, walking, or running. For example, carrying while crouching
requires the controller to maintain support of the box while realizing a
substantially different whole-body configuration. The resulting rollouts add
diverse compositional behavior to the downstream planner data.

For an interaction $I$, let $z^N\in\mathcal Z$ denote the native BFM behavior
to be composed with it. We freeze the BFM and learn an adapter
$A_{\psi_I}$ that outputs an executable latent,
\begin{equation}
    \hat z_t
    =
    \Pi_{\mathcal Z}
    \left(z^N+A_{\psi_I}^{\mathrm{res}}(o_t,z^N)\right),
    \quad
    a_t=\pibfm(s_t,\hat z_t).
    \label{eq:app_adapter}
\end{equation}
The residual parameterization lets the adapter modify the native command only
as needed to satisfy the interaction. The deployable adapter uses causal robot
inputs and the requested native latent; privileged object state and interaction
errors may be used by the critic and reward computation during simulation
training.

We optimize the adapter with
\begin{equation}
    r_t
    =
    r_t^{I}
    +
    \lambda_{\mathrm{SF}} B(s_t)^\top z^N,
    \label{eq:app_sf_reward}
\end{equation}
where $r_t^{I}$ preserves the interaction and the SF term encourages the
rollout to remain aligned with the requested native behavior. For fixed
$z^N$,
\begin{equation}
    \mathbb E\!\left[
        \sum_{t=0}^{\infty}\gamma^t B(s_t)^\top z^N
    \right]
    =
    \mathbb E\!\left[
        \sum_{t=0}^{\infty}\gamma^t B(s_t)
    \right]^\top z^N,
    \label{eq:app_sf_identity}
\end{equation}
so the preservation term evaluates the discounted state occupancy along the
behavioral direction selected by $z^N$. The two terms therefore play distinct
roles: $r_t^I$ keeps the interaction valid, while the SF term determines which
native behavior is composed with that interaction.

For the carrying compositions used in our experiments, $r_t^I$ contains
upper-body pose, object-position, object-orientation, and contact terms. The
three tracking terms are
\begin{align}
    r_{\mathrm{pose}}&=\exp(-e_q/0.25^2),\\
    r_{\mathrm{pos}}&=\exp(-\|e_p\|_2^2/0.10^2),\\
    r_{\mathrm{ori}}&=\exp(-\theta_{\mathrm{obj}}^2/0.35^2),
\end{align}
with weights $1$, $2$, and $1$, respectively; the contact term has weight $1$.
Here $e_q$ is the mean squared upper-body joint-position error, $e_p$ is the
object-position error in the torso frame, and $\theta_{\mathrm{obj}}$ is the
object-orientation discrepancy in that frame. The SF term has weight $0.1$,
the latent-smoothness penalty has weight $-0.5$, and object-drop termination
contributes $-100$. These interaction terms are shared across the composed
variants; behavioral diversity is introduced by changing $z^N$.

Because both $B(s_t)$ and $z^N$ lie in the radius $\sqrt{d_z}$ sphere,
\begin{equation}
    \frac{B(s_t)^\top z^N}{d_z}
    =
    \cos\angle\!\left(B(s_t),z^N\right).
\end{equation}
The normalized and unnormalized forms therefore differ only by a constant
scale absorbed into $\lambda_{\mathrm{SF}}$. Importantly, the preservation
target is the requested native behavior $z^N$, not the adapted latent
$\hat z_t$.

The adapter actor and critic use two residual feed-forward blocks of width 512
with feed-forward width 2048, GELU activations, and observation normalization.
We additionally regularize rapid latent changes with the normalized cost
\begin{equation}
    c_{\mathrm{smooth},t}
    =
    \frac{\|\hat z_t-\hat z_{t-1}\|_2^2}{2d_z},
\end{equation}
which equals one minus cosine similarity on $\mathcal Z$.

\section{Action-Aligned Latent Policy Projection}
\label{app:lpp}

After each expert is acquired, \ac{lpp} maps its local closed-loop responses into
the executable BFM space. This section follows the same order as the main
method: first establish the shared action convention, then define the projector
and short-horizon latent execution, the action/semantic objectives, and finally
the on-policy aggregation used to obtain robust projected rollouts.

\subsection{Common Action Space}

Experts may use different normalized action conventions, so action alignment is
performed only after both expert and BFM outputs are converted to the same
physical body joint-target space. The frozen BFM actor produces
$a_t=\pibfm(s_t,z_t)\in\mathbb R^{29}$. Under the BFM control profile,
\begin{equation}
    q_t^{\mathrm{des}}
    =
    q^{\mathrm{default}}
    +
    0.25\,r_q\odot
    \operatorname{clip}(6a_t,-6,6),
    \label{eq:app_action_conversion}
\end{equation}
where $r_q$ contains the per-joint action scales. Joint order, default pose,
and scale vectors are kept consistent across expert conversion, projection,
projected-rollout collection, and execution. The resulting targets are
realized by the robot PD controller with the gains and actuator limits of the
selected control profile.

\subsection{Projector and Latent Execution}

Each expert is projected independently. At update time $t_k$, a projector
$P_{\psi_i}$ maps the expert observation to BFM latent,
\begin{equation}
    z_k=P_{\psi_i}(o_{t_k}^{E_i}),
    \quad z_k\in\mathcal Z.
    \label{eq:app_projector}
\end{equation}
The projector is a pre-normalized residual MLP consisting of an input LayerNorm
and linear projection, residual feed-forward blocks, an output LayerNorm, and a
latent head followed by $\Pi_{\mathcal Z}$. A residual block is
\begin{align}
    \tilde h^\ell
    &=
    \operatorname{GELU}
    \left(W_1^\ell\operatorname{LN}(h^\ell)+b_1^\ell\right),\\
    h^{\ell+1}
    &=h^\ell+W_2^\ell\tilde h^\ell+b_2^\ell.
\end{align}
The hidden width is 1024 with feed-forward expansion ratio four. The
motion-based projector uses four residual blocks and the door projector uses
two.

Projection updates occur every five body-control steps, corresponding to a
$0.1$\,s interval and a $10$\,Hz projector. Between two projector updates, the
latent command is interpolated on the sphere,
\begin{equation}
    z_{t_k+h}
    =
    \operatorname{SLERP}
    \left(z_{k-1},z_k,\alpha_h\right),
    \quad h=0,\ldots,H_P-1,
    \label{eq:app_slerp_execution}
\end{equation}
where $H_P=5$. For endpoints $a,b\in\mathcal Z$ with angular separation
$\vartheta$,
\begin{equation}
    \operatorname{SLERP}(a,b,\alpha)
    =
    \frac{\sin((1-\alpha)\vartheta)}{\sin\vartheta}a
    +
    \frac{\sin(\alpha\vartheta)}{\sin\vartheta}b.
\end{equation}
The cosine used to compute $\vartheta$ is clamped for numerical stability.

\subsection{Action Alignment and Semantic Regularization}

Let $T_B$ and $T_i$ convert the BFM and expert outputs to the same physical
body joint-target space. Over the short execution interval beginning at
$t_k$, we use
\begin{align}
    e_h^a
    &=
    T_B\!\left(\pibfm(s_{t_k+h},z_{t_k+h})\right)
    -
    T_i\!\left(\pi_i^E(o_{t_k+h}^{E_i})\right),\\
    \mathcal L_{\mathrm{act}}
    &=
    \mathbb E_{P_{\psi_i}}
    \left[
    \frac{\sum_{h=0}^{H_P-1}m_h\|e_h^a\|_2^2}
         {29\sum_{h=0}^{H_P-1}m_h}
    \right],
    \label{eq:app_target_loss}
\end{align}
where $m_h$ masks samples beyond an episode boundary. Comparing physical joint
targets prevents normalized-action conventions of different experts from
changing the meaning of the alignment loss.

Action matching alone can leave the latent underdetermined. We therefore
anchor the predicted latent to the native BFM representation of the same local
state,
\begin{equation}
    \mathcal L_{\mathrm{sem}}
    =
    \mathbb E
    \left[
        \|z_k-B(s_{t_k})\|_2^2
    \right].
    \label{eq:app_semantic_loss}
\end{equation}
For two vectors on $\mathcal Z$,
\begin{equation}
    \|z-B(s)\|_2^2
    =
    2d_z
    \left(1-\frac{z^\top B(s)}{d_z}\right),
    \label{eq:app_semantic_equiv}
\end{equation}
so squared distance and cosine gap are equivalent up to a constant factor. The
full objective is
\begin{equation}
    \mathcal L_{\mathrm{proj}}
    =
    \mathcal L_{\mathrm{act}}
    +
    \lambda_{\mathrm{sem}}\mathcal L_{\mathrm{sem}}.
    \label{eq:app_proj_full}
\end{equation}
The action term identifies commands by the expert's required local control
response, whereas the semantic term resolves action-equivalent solutions in a
way that remains coherent with the native BFM geometry. Training details in~\cref{tab:app_projector_train}.

For skills with dexterous hands, a separate projector head regresses the
expert's physical hand-joint targets. Its loss is evaluated only when valid
hand labels are available. The sphere constraint applies only to the body
latent; hand outputs retain their physical joint-target parameterization.

\subsection{On-Policy Projection and Data Aggregation}

Projection is trained with single-skill on-policy data aggregation. The current
projector is executed through the frozen BFM under randomized task
initializations. At the states it visits, we query the corresponding expert
for body actions and record the BFM inputs, expert observations, and
reset masks. Short execution windows are then replayed through the frozen BFM
actor to evaluate \eqref{eq:app_target_loss} and
\eqref{eq:app_semantic_loss}. The projector is updated and new rollouts are
collected with the updated policy. Gradients pass through the frozen BFM actor
to the predicted latent, but no BFM parameter is changed and no simulator
gradient is required. This DAgger-style procedure~\cite{ross2011dagger}
exposes the projector to corrective states that are absent from nominal expert
trajectories.

\begin{table}[t]
    \centering
    \caption{Latent-projector optimization parameters.}
    \label{tab:app_projector_train}
    \small
    \begin{tabular*}{\columnwidth}{
    @{\hspace{\tabcolsep}}
    l
    @{\extracolsep{\fill}}
    c
    @{\hspace{\tabcolsep}}
    }
        \toprule
        Parameter & Value \\
        \midrule
        Optimizer & AdamW \\
        Learning rate & $3\times10^{-4}$ \\
        Weight decay & $10^{-4}$ \\
        Gradient-norm limit & 1 \\
        Minibatches / rollout & 4 \\
        Optimization passes / rollout & 5 \\
        Semantic-loss weight $\lambda_{\mathrm{sem}}$ & 0.1 \\
        Body-control steps / latent interval $H_P$ & 5 \\
        Executed latent noise std. & 0.1 \\
        \bottomrule
    \end{tabular*}
\end{table}

Exploration noise is applied to the executed latent before projection back to
$\mathcal Z$ and is never added to the expert action label. Once projection is
complete, the original expert is no longer required for projected execution or
planner-data collection.

The projected latent should be interpreted locally: it is identified by the
control response induced by the frozen BFM on states visited during projection,
rather than by assuming that the expert and BFM policies are globally identical.
This is why projected closed-loop rollouts, rather than nominal expert
trajectories alone, are used as the supervision for the unified planner.

\section{Unified Latent Flow Planning}
\label{app:planner}

\begin{table*}[!tb]
    \centering
    \caption{Architecture of the unified latent flow planner.}
    \label{tab:app_planner_arch}
    \small
    \setlength{\tabcolsep}{4pt}
    \begin{tabular*}{\textwidth}{
        @{}
        l
        @{\extracolsep{\fill}}
        c
        l
        @{\extracolsep{\fill}}
        c
        @{}
    }
        \toprule
        Component or parameter & Configuration
        & Component or parameter & Configuration \\
        \midrule
        Executable body latent
        & $d_z=256$
        & Transformer width / latent blocks
        & 1024 / 16 \\

        Prediction horizon
        & $1$\,s
        & Attention heads / per-head width
        & 8 / 128 \\

        Body / optional hand state
        & 58 / 14
        & Feed-forward expansion / activation
        & 4 / GELU \\

        Optional hand target
        & 14 per latent step
        & Attention / feed-forward dropout
        & 0 / 0.2 \\

        Depth image / patch shape
        & $144\times256$ / $12\times16$
        & Conditioning registers
        & 8 \\

        Depth range / spatial tokens
        & $0.12$--$3.0$\,m / 192
        & Flow-time embedding scale
        & 1000 \\

        Depth-encoder blocks / registers
        & 2 / 8
        & Pooled language embedding
        & 768 (frozen T5-base) \\

        Waypoint input
        & 10, at $0.5$ and $1.0$\,s
        & Steering blocks / output tokens
        & 2 / 8 \\
        \bottomrule
    \end{tabular*}
\end{table*}

Once each expert has been projected, its closed-loop latent rollouts can be
combined because they now share the same executable representation. We train
the unified planner entirely offline on these projected trajectories, matching
the data construction, conditioning variables, and flow objective. Details of the planner network are shown in~\cref{tab:app_planner_arch}.

\subsection{Projected Rollout Dataset and Prediction Target}

For each projected skill, we collect 20K closed-loop episodes and aggregate
them for offline planner training. Each record contains onboard depth,
proprioception, the high-level condition, and the action-aligned latent actually
executed by the projected policy. Replacing these targets with $B(s)$ would
change the supervision from projected control responses to behavioral-state
encoding and is therefore not equivalent to our data construction.

The planner predicts a one-second sequence of future BFM commands. We use one
latent target every $0.1$\,s, corresponding to five $50$\,Hz BFM control
steps, and predict $H=10$ latent commands per sequence. Matching the trajectory
notation used in the main paper,
\begin{equation}
    Z_t
    =
    [z_t,z_{t+1},\ldots,z_{t+H-1}]
    \in\mathbb R^{H\times d_z},
    \quad
    H=10,
    \label{eq:app_planner_target}
\end{equation}
Each future latent is represented by one temporal token in the transformer.
For hand-bearing data, a separate target
$Y_t\in\mathbb R^{H\times14}$ stores physical hand-joint targets in a fixed
hand-major ordering. Hand targets are normalized with their own statistics and
are not projected onto $\mathcal Z$.

\subsection{Observation and Steering Conditioning}

Unlike the task-specific experts, the unified planner uses only onboard
observations. Its proprioceptive input is
$p_t=[q_t;\dot q_t]\in\mathbb R^{58}$; for hand-bearing samples, the measured
14 hand-joint positions are included through a separate projection. Continuous
features are standardized coordinate-wise using fixed training-set statistics,
and the normalization statistics and joint ordering are stored with the model.

\paragraph{Depth}
The planner processes a metric egocentric depth image of size
$144\times256$, clipped to the valid range $0.12$--$3.0$\,m. A strided
convolution with $12\times16$ kernel and stride produces a $12\times16$ grid
of 192 spatial tokens. Learned spatial embeddings are added before two
self-attention blocks. Eight learned register tokens participate in the depth
encoder and are discarded afterward, leaving the 192 spatial tokens as
conditioning memory.

\paragraph{Proprioception}
The 58-dimensional joint state is standardized coordinate-wise and linearly
embedded to the model width. For coordinate $j$,
$\bar p_j=(p_j-\mu_j)/\max(\sigma_j,\epsilon)$. When hands are enabled, their
measured joint positions are standardized separately and added through an
additional projection.

\paragraph{Language}
Task instructions are encoded by a frozen T5-base text encoder. Masked pooling
produces a 768-dimensional sentence embedding, followed by LayerNorm and a
two-layer SiLU MLP that maps it to the planner width. The text encoder is not
updated by the planner objective.

\paragraph{Waypoint}
Waypoint supervision is derived from future root poses. At horizons
$\delta\in\{0.5,1.0\}$\,s, we use
\begin{equation}
    u_t^\delta
    =
    [
       \Delta x_t^\delta,
       \Delta y_t^\delta,
       \sin\Delta\psi_t^\delta,
       \cos\Delta\psi_t^\delta,
       h_{t+\delta}
    ],
    \label{eq:app_waypoint}
\end{equation}
where displacement is expressed in the current heading frame,
$\Delta\psi$ is relative yaw, and $h$ is future pelvis height. Concatenating
the two horizons gives a ten-dimensional waypoint condition. Missing waypoint
conditions are masked rather than represented by a zero-valued physical
command.

\paragraph{Steering representation}
Language and waypoint embeddings are converted to a small fixed set of steering
tokens by a self-attention encoder. The latent transformer cross-attends to
these steering tokens, the 192 spatial depth tokens, one proprioceptive state
token, and eight learned conditioning-register tokens. Unavailable language or
waypoint modalities are masked explicitly.

\subsection{Conditional Latent Transformer}

Each noisy future latent is linearly embedded and combined with a learned
temporal-position embedding. If hand targets are present, a separate
projection of the noisy hand target is added to the corresponding temporal
token. Flow time is sinusoidally embedded and passed through an MLP. Each
transformer block contains temporal self-attention, cross-attention to the
observation/steering memory, and a feed-forward MLP. Adaptive LayerNorm
modulation supplies shift, scale, and residual-gate vectors from the flow-time
condition. For example,
\begin{equation}
    h'
    =
    h+g_\tau\odot
    \operatorname{Attn}
    \left((1+s_\tau)\odot\operatorname{LN}(h)+b_\tau\right).
\end{equation}
Separate linear heads decode the clean body latents and optional hand targets.

\subsection{Offline Flow-Matching Objective}
\label{app:flow}

Let $\xi_t$ denote the encoded onboard observations and available steering
conditions. We sample $x_0\sim\mathcal N(0,I)$ and
$\tau\sim\mathcal U(0,1)$, and construct
\begin{equation}
    x_\tau=(1-\tau)x_0+\tau x_1.
\end{equation}
The network predicts the clean endpoint
$f_\eta(x_\tau,\tau,\xi_t)$ and is trained with
\begin{equation}
    \mathcal L_{\mathrm{body}}
    =
    \mathbb E
    \left[
       \frac{1}{Hd_z}
       \|f_\eta(x_\tau,\tau,\xi_t)-x_1\|_2^2
    \right].
    \label{eq:app_body_flow}
\end{equation}
The clean latent targets lie on $\mathcal Z$, but the Gaussian interpolation and
raw endpoint prediction are carried out in Euclidean coordinates. Sphere
projection is applied only when sampled outputs are converted to executable BFM
commands.

An analogous Gaussian interpolation is applied to the hand targets when available. The complete loss is
\begin{equation}
    \mathcal L_{\mathrm{planner}}
    =
    \mathcal L_{\mathrm{body}}
    +
    \lambda_{\mathrm{hand}}\mathcal L_{\mathrm{hand}},
    \label{eq:app_joint_flow}
\end{equation}
where $\mathcal L_{\mathrm{hand}}$ is evaluated only on valid hand samples. No
decoded body-action auxiliary loss is used in the reported planner training;
action alignment is already incorporated when constructing the projected latent
supervision.

An endpoint predictor induces the velocity field
\begin{equation}
    v_\eta(x_\tau,\tau,\xi_t)
    =
    \frac{f_\eta(x_\tau,\tau,\xi_t)-x_\tau}{1-\tau},
    \quad \tau<1.
    \label{eq:app_endpoint_velocity}
\end{equation}
For a numerical grid $0=\tau_0<\cdots<\tau_N=1$, Euler integration uses
\begin{equation}
    x_{j+1}
    =
    x_j+(\tau_{j+1}-\tau_j)v_\eta(x_j,\tau_j,\xi_t).
\end{equation}
Heun integration uses the corresponding predictor-corrector update. After
sampling, every body latent is independently projected by $\Pi_{\mathcal Z}$,
while hand outputs are denormalized using their own statistics.

\subsection{Planner Optimization}

\begin{table}[!tb]
    \centering
    \caption{Offline planner optimization parameters.}
    \label{tab:app_flow_train}
    \small
    \begin{tabular*}{\columnwidth}{
    @{\hspace{\tabcolsep}}
    l
    @{\extracolsep{\fill}}
    c
    @{\hspace{\tabcolsep}}
    }
        \toprule
        Parameter & Value \\
        \midrule
        Optimizer & AdamW \\
        Learning rate & $2\times10^{-4}$ \\
        Weight decay & $10^{-2}$ \\
        Warm-up updates & 100 \\
        Post-warm-up schedule & Constant \\
        Batch size & 256 \\
        Training epochs & 500 \\
        Gradient-norm limit & 1 \\
        Body endpoint-loss weight & 1 \\
        Hand endpoint-loss weight & 1 \\
        Decoded body-action loss weight & 0 \\
        Default windows / worker & 16 \\
        Validation fraction & 0.2 \\
        Flow-time distribution & $\mathcal U(0,1)$ \\
        Noise distribution & $\mathcal N(0,I)$ \\
        \bottomrule
    \end{tabular*}
\end{table}

Shown in~\cref{tab:app_flow_train}, training windows are sampled within episode boundaries, and train/validation
partitions are defined at the episode level. The model checkpoint stores the
architecture, optimizer state, proprioceptive normalization, and hand-target
normalization. In the exported configuration, state-condition dropout is
$0.2$ and waypoint-condition dropout is $0.1$; waypoint tokens are omitted for
episodes without waypoint supervision.

\subsection{Receding-Horizon Execution}

At deployment, the planner executes the leading segment of the sampled latent
sequence and then replans from the new onboard observation. Consecutive latent
updates are connected by SLERP to supply the BFM with a command at each $50$\,Hz
body-control step. The BFM actor therefore retains high-frequency
state-dependent feedback while the planner operates at a lower rate. Hand
joint targets follow the same receding-horizon schedule as the corresponding
body-latent sequence.

\begin{table*}[!tb]
    \centering
    \caption{Simulation evaluation variations.}
    \label{tab:app_eval_variations}
    \small
    \setlength{\tabcolsep}{6pt}
    \begin{tabularx}{\textwidth}{l l X}
        \toprule
        Task & Variation & Evaluation range \\
        \midrule
        Box pickup
        & Box geometry, placement, and mass
        & Base size $20\times30\times40$\,cm, independently scaled by
        $s_x,s_y\in[0.5,1.5]$ and $s_z\in[0.5,1.0]$,
        yielding sizes from $10\times15\times20$\,cm to
        $30\times45\times40$\,cm.
        Boxes are placed either on the ground or on supports $0.25$--$0.50$\,m high,
        with mass sampled from $0.5$--$1.5$\,kg. \\
        
        Door opening
        & Door geometry and handle placement
        & Door width $0.8$--$1.1$\,m and height $1.9$--$2.2$\,m.
        Handle height is $0.85$--$0.95$\,m above the ground, with lateral inset
        $0.08$--$0.15$\,m. \\
        
        Object grasping
        & Object identity and shape
        & $128$ distinct object instances are evaluated with four takes per object
        ($512$ episodes total), selected through teacher-feasibility screening from
        $685$ object instances and $2{,}991$ usable demonstrations. \\
        \bottomrule
    \end{tabularx}
\end{table*}

\section{Evaluation and Real-World Deployment}

\subsection{Simulation Evaluation Protocol}

We evaluate each task over diverse object or environment variations in IsaacSim.
For pickup and door opening, the \emph{w/ appr.} setting initializes the humanoid approximately $0.5$--$1.5$\,m from the interaction target and requires it to approach before execution, whereas \emph{w/o appr.} initializes the humanoid directly in front of the target.
Pickup succeeds if the box is lifted by at least $0.3$\,m and maintained for at least $1$\,s; grasping succeeds if the object is lifted by at least $0.1$\,m and maintained for at least $1$\,s.
Door opening succeeds when the door is opened and the humanoid passes through the doorway.
Reported $\pm$ values are computed from repeated evaluations with different random seeds to account for simulation stochasticity. Detailed object variation configurations used in evaluation are shown in~\cref{tab:app_eval_variations}.

\subsection{Real-World Execution}

The real-world system uses onboard RealSense depth and proprioception.
Both the unified planner and the frozen BFM run on a single NVIDIA RTX PRO 5000 Blackwell laptop.
The TensorRT planner replans at $5$\,Hz using $10$ Heun integration steps, with approximately $20$\,ms inference time per planner call, while the BFM runs at $50$\,Hz with approximately $1$\,ms per forward pass.

We use \ac{rtc} during deployment to avoid discontinuities between consecutively sampled latent trajectories.
At each replanning step, the flow planner predicts a chunk of future latent commands while the robot continues executing the current chunk.
The newly generated chunk overlaps temporally with the remaining portion of the previous prediction, and \ac{rtc} conditions the overlapping prefix of the new sample toward the previously committed trajectory during flow integration.
This preserves continuity across replanning boundaries while still allowing later latent commands in the chunk to adapt to the latest onboard observation.
In this way, \ac{rtc} decouples the planner inference latency from continuous robot execution and avoids abrupt latent changes caused by independently sampling each plan.

\end{document}